\documentclass[10pt,twocolumn,letterpaper]{article}

\usepackage{iccv}              % To produce the CAMERA-READY version
\definecolor{iccvblue}{rgb}{0.21,0.49,0.74}
\usepackage[pagebackref,breaklinks,colorlinks,allcolors=iccvblue]{hyperref}

\def\paperID{6873} % *** Enter the Paper ID here
\def\confName{ICCV}
\def\confYear{2025}

\usepackage{multirow}
\usepackage{graphicx}
\usepackage{amsmath}
\usepackage{amssymb}
\usepackage{mathtools, nccmath}
\usepackage{amsthm}
\usepackage{pifont}

\usepackage{algorithm, algorithmic}
\usepackage{wrapfig, lipsum, booktabs}
\usepackage{colortbl}
\usepackage{comment}
\usepackage{marvosym}      % colors
\usepackage{subcaption}
 \usepackage{enumitem}
 \usepackage{xcolor} % 引入xcolor包

\usepackage{bm}
\usepackage{mathtools}
\usepackage{subcaption}

\definecolor{mygray}{RGB}{127,127,127}
\definecolor{mygreen}{RGB}{93,174,86}

\usepackage{tcolorbox}
\tcbuselibrary{skins, breakable, theorems}

\makeatletter
\newcommand{\thickhline}{%
	\noalign {\ifnum 0=`}\fi \hrule height 1pt
	\futurelet \reserved@a \@xhline
}
\makeatother

\usepackage{caption}
\usepackage{caption}
\newcommand\blfootnote[1]{%
  \begingroup
  \renewcommand\thefootnote{}\footnote{#1}%
  \addtocounter{footnote}{-1}%
  \endgroup
}

\title{RAGNet: Large-scale Reasoning-based Affordance Segmentation Benchmark towards General Grasping}
\author{
Dongming Wu$^{1}$, Yanping Fu$^{2\star}$, Saike Huang$^{3}$, Yingfei Liu$^{3\dagger}$, Fan Jia$^{3}$, Nian Liu$^{4}$, Feng Dai$^{2}$, \\
Tiancai Wang$^{3}$, Rao Muhammad Anwer$^{4}$, Fahad Shahbaz Khan$^{4}$, Jianbing Shen$^{5\ddagger}$\\
{\footnotesize
$^1$ The Chinese University of Hong Kong,
$^2$ Institute of Computing Technology, Chinese Academy of Sciences, 
 } \\
{\footnotesize
$^3$ Dexmal,
$^4$ Mohamed bin Zayed University of Artificial Intelligence,
$^5$ SKL-IOTSC, CIS, University of Macau 
}}
\begin{document}
% \maketitle

\twocolumn[{
\renewcommand\twocolumn[1][]{#1}%
\maketitle
 
\begin{figure}[H]
\small
\hsize=\textwidth
\centering
    \vspace{-1cm}
    \includegraphics[scale=0.5]{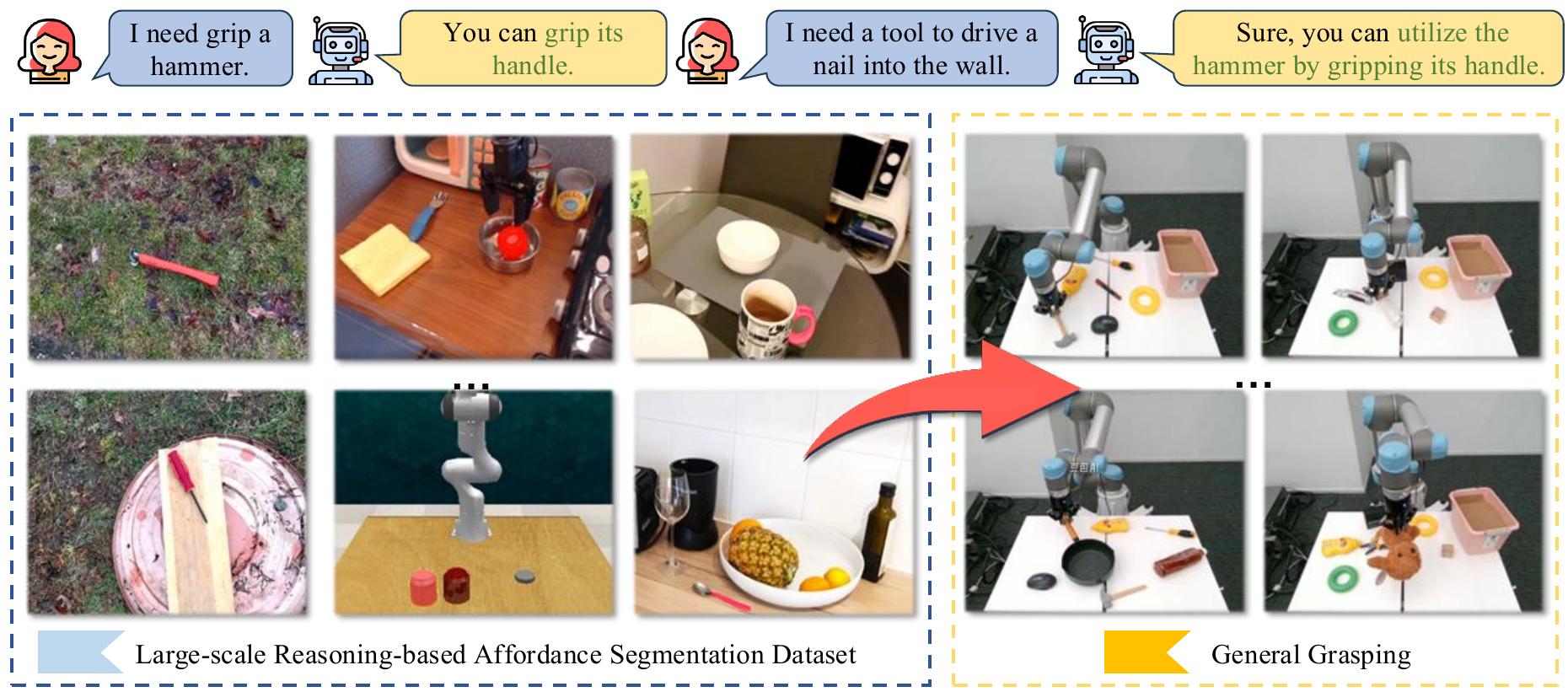}
    \vspace{-10pt}
    \caption{\textbf{Large-scale benchmark for reasoning-based affordance segmentation}, which sources from various embodied domains: wild, robot, ego-centric indoor, and simulation.  
    %With training on the massive high-quality data, our model AffordaceNet shows impressive zero-shot or out-of-domain generalization ability, towards powerful general grasping ability.
    By leveraging the extensive high-quality dataset for training, our model \textit{AffordanceNet} exhibits remarkable open-world generalization capabilities, steering further towards robust general-purpose object grasping.
    %The objective of our research is to investigate the expansion of these affordance segmentation datasets to enhance their applicability for generalizing and manipulating in an open-world context.
    }
    \label{fig:dataset}
\end{figure}
}]

\blfootnote{
$\star$ The work is done during the internship at Dexmal.
$\dagger$ Project lead. 
$\ddagger$ Corresponding author. 
This work was supported in part by the Science and Technology Development Fund of Macau SAR (FDCT) under grants 0102/2023/RIA2 and 0154/2022/A3 and 001/2024/SKL, the Jiangyin Hi-tech Industrial Development Zone under the Taihu Innovation Scheme (EF2025-00003-SKL-IOTSC), and the University of Macau SRG2022-00023-IOTSC grant.
}

\begin{abstract}
General robotic grasping systems require accurate object affordance perception in diverse open-world scenarios following human instructions. However, current studies suffer from the problem of lacking reasoning-based large-scale affordance prediction data, leading to considerable concern about open-world effectiveness. To address this limitation, we build a large-scale grasping-oriented affordance segmentation benchmark with human-like instructions, named RAGNet. It contains 273k images, 180 categories, and 26k reasoning instructions. The images cover diverse embodied data domains, such as wild, robot, ego-centric, and even simulation data. They are carefully annotated with an affordance map, while the difficulty of language instructions is largely increased by removing their category name and only providing functional descriptions. Furthermore, we propose a comprehensive affordance-based grasping framework, named AffordanceNet, which consists of a VLM pre-trained on our massive affordance data and a grasping network that conditions an affordance map to grasp the target. Extensive experiments on affordance segmentation benchmarks and real-robot manipulation tasks show that our model has a powerful open-world generalization ability. Our data and code  is available at \href{https://github.com/wudongming97/AffordanceNet}{this link}.
\end{abstract}    
\section{Introduction}
\label{sec:intro}

\begin{table*}[t]
\centering
\small
\resizebox{0.8\linewidth}{!}{
		\setlength\tabcolsep{4pt}
		\renewcommand\arraystretch{1}
    \begin{tabular}{l|cc|cccc|cc}
    \hline \thickhline
    Dataset & Images & Categories & Wild & Robot & Ego-centric & Simulation & Reasoning & Output\\
    \hline
    UMD~\cite{umd} \textcolor{gray}{\footnotesize{ICRA2015}} &10k  & 17 & & \checkmark&&&\checkmark&Seg\\
    % OPRA~\cite{fang2018demo2vec} \textcolor{gray}{\footnotesize{CVPR2018}} &  (11k) & - & && \checkmark &&& Seg\\
    AGD20k~\cite{agd20k} \textcolor{gray}{\footnotesize{CVPR2020}}& 20k & 50  & \checkmark & &&&&Seg\\
    HANDAL~\cite{handal} \textcolor{gray}{\footnotesize{IROS2023}} & 200k & 17 & \checkmark & & & & & Seg/Box \\
    AED~\cite{li2024learning}\textcolor{gray}{\footnotesize{Arxiv2024}}& - &21  & & \checkmark && \checkmark & &Seg \\
    3DOI~\cite{3doi}\textcolor{gray}{\footnotesize{ICCV2023}} & 10k & - & \checkmark & & \checkmark & && Seg \\
    AffordanceLLM~\cite{affordancellm}\textcolor{gray}{\footnotesize{CVPR2024}} & 20k & -  & \checkmark & &&&&Seg\\
    ManipVQA~\cite{manipvqa}~\textcolor{gray}{\footnotesize{IROS2024}} & 84k & - & \checkmark & & \checkmark & & \checkmark & Seg \\
    RAGNet (Ours) & 273k & 180 & \checkmark & \checkmark & \checkmark & \checkmark & \checkmark & Seg\\
    \hline
    \thickhline
    \end{tabular}
    }
    \caption{\textbf{Comparisons between previous affordance data and our collection.} ``(11k)'' represents the number of video clips. ``-'' means unavailable data. ``Reasoning'' refers to reasoning instructions. }
    \vspace{-12pt}
    \label{tbl:data}
\end{table*}

% Affordance segmentation is an important task for robotic manipulation, which requires grounding the object region that the robot can grasp.
% The concept of affordance is defined as those actions or behaviors afforded due to the physical structure and design of an object (e.g., a button affords to press, a drawer knob affords to pull).

%Machine perception is a foundational research topic that plays an important role in various downstream embodied tasks, such as autonomous driving~\cite{uniad,nuscenes} and robotics navigation and manipulation.
Affordance prediction is a foundational research topic that significantly contributes to diverse practical applications, including robotic manipulation~\cite{huang2023voxposer,shridhar2022cliport,huang2024copa,huang2024rekep,nasiriany2024pivot} and human-object interaction~\cite{shan2020understanding,kim2021hotr,bahl2023affordances,jian2023affordpose,wang2024move,wang2024interactive}. 
It requires comprehensively understanding the geometry and function of an object (\eg, a wok affords to hold, a microwave door affords to pull) for further detecting graspable regions.
Nonetheless, this task faces two additional primary difficulties in the realm of open-world generalization concerning vision and language instructions: 
1) The capability to generalize in a broad range of unseen object categories and image domains. 
%1) The capability to generalize in open-world scenarios, that is, generalizing on unseen objects and image domains. 
%2) The necessity for algorithms to thoroughly grasp the physical properties and functions of objects to precisely identify regions suitable for robotic grasping. 
2) The alignment of commands that mimic human-like high-level instructions.

Initial studies often specialize in particular domains, as depicted in Table~\ref{tbl:data}.
For instance, UMD~\cite{umd} offers 10,000 RGB-D image pairs from three cluttered scenes with pixel-level affordance annotations.
This dataset is categorized as \textit{robot data}, characterized by a distinct operational table and a relatively stable background.
Despite the abundance of public robot datasets~\cite{openx,brohan2022rt,walke2023bridgedata}, they often lack meticulous fine-grained labeling.
\textit{Ego-centric data}, which is captured from a first-person or egocentric viewpoint, is the most prevalent type of data~\cite{ego4d,damen2022rescaling}.
Nevertheless, this data is often collected in indoor kitchens and workshops, resulting in a lack of diversity in the data category distribution.
\textit{Wild data}, another prevalent type, is gathered in various situations, which usually cover diverse categories~\cite{agd20k,paco,vlpart}. However, certain categories within this data, like bicycles, motorcycles, sofas and \etc, are not suitable for robotic manipulation.
%In addition, these data have a great domain gap under various conditions (like background, lights, and object categories).
In addition, these data often fail to generalize to unseen domains and novel objects for affordance prediction.

%like robot data, wild data, or simulation.
To enhance open-world generalization, Vision Language Model (VLM) with massive image-text training~\cite{llava,li2023blip} has become an important increment in various visual prediction tasks~\cite{lisa,glamm}.
%With the recent advancement of Vision Language Models (VLMs) with massive image-text pre-training~\cite{llava,li2023blip}, world knowledge becomes an important increment for various downstream open-world generalization and reasoning-based tasks~\cite{lisa,glamm}.
Inspired by this, VLM also has an increasing interest in affordance prediction along with complex reasoning, and several recent works are attempting to predict affordance regions~\cite{affordancellm,manipvqa,li2024one,li2024manipllm,yu2024uniaff,xu2024naturalvlm}.
For example, ManipVQA~\cite{manipvqa} proposes  tool detection, affordance recognition, and a
broader understanding of physical concepts in a unified framework.  
However, these works employ a fixed format of language prompt and a limited data scale, which brings some concerns about open-world generalization and complex reasoning.
Overall, current efforts are either constrained by insufficiently diverse datasets or by the absence of effective reasoning mechanisms, which hampers the ability to perform open-world grasping.

\begin{figure*}[t]
% \hsize=\textwidth
\centering
    % \vspace{-1cm}
    \includegraphics[width=0.9\linewidth]{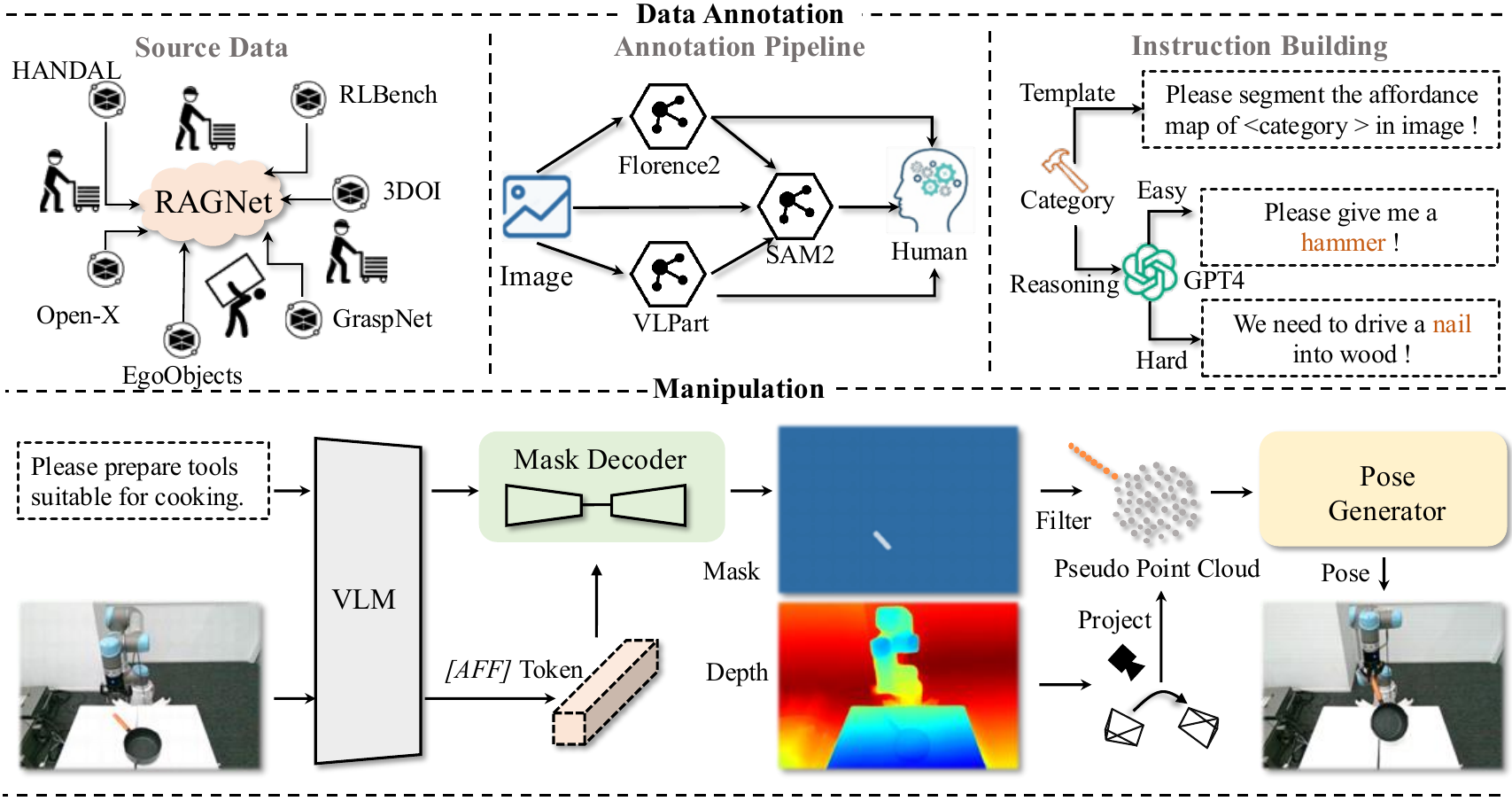}
    \vspace{-10pt}
    \caption{\textbf{
Overview of our data annotation pipeline and manipulation model.}  We collect data from several public datasets, including HANDAL, GraspNet, Open-X, etc. A variety of models and manual annotation are utilized to annotate affordance masks. Subsequently, we refine textual instructions using templates and GPT4 to emulate human-like commands. For grasp operations, the VLM model is employed to identify affordance regions, which are then converted into the required grasp poses by integrating depth information.
}
    \vspace{-4pt}
    \label{fig:framework}
\end{figure*}

To address these challenges, we first build large-scale affordance segmentation data from various image sources, including wild, robot, ego-centric, and even simulated data, named \textbf{RAGNet}.
It has 273k images and 180 categories.
We design a set of affordance annotation tools for labeling the regions of objects that are amenable to grasping.
Furthermore, we leverage Large Language Models (LLMs) to generate a vast array of reasoning-based instructions, totaling 26k distinct expressions.
Here, we create two types of reasoning-based instructions beyond the template-based.
One includes the name of the object and the other omits it. Take a knife as an example: ``Please provide a knife" versus ``I want something to slice the bread". This approach closely mirrors real-life human interactions. 
In summary, we have constructed a massive-scale database with domains, object categories, and complex reasoning instructions.

Furthermore, we introduce an affordance-based grasping framework, titled \textbf{AffordanceNet}.
It presents a deployable and general grasping pipeline, which stands out from prior MLLM-based affordance prediction methods~\cite{affordancellm,manipvqa} that have not yet demonstrated real-robot deployment.
Our model includes two crucial components, AffordanceVLM and Pose Generator.
The AffordanceVLM transforms RGB images and human instruction into an accurate affordance map, while the pose generator uses the 2D affordance with a depth image to produce 3D grasper pose.
%After the training of the Vision-Language Model (VLM) on this dataset, we conduct comprehensive experiments to assess its open-world generalization and reasoning capabilities. 
In this work, we conduct extensive experiments to assess its open-world generalization and reasoning capabilities. 
First, we create two distinct validation datasets to assess open-world generalization: one for zero-shot categoty recognition and another for out-of-domain affordance prediction. 
Second, to test reasoning ability, we implement an affordance segmentation validation based on instructions that do not contain any target category names.
%Thirdly, we conduct a set of real-robot experiments in new zero-shot situations.
Thirdly, we carry out a range of close-loop real-robot grasping tasks in an entirely out-of-domain setting.
Last but not least, we test several representative simulation tasks from RLBench.
%for better close-loop reproducibility.
All experiments show that our proposed method has great generalization and reasoning ability.
In summary, our contributions are three-fold:
\begin{itemize}
    \item 
    %We explore a reasoning-based affordance segmentation at scale. Its core idea is to evaluate the open-world generalization and reasoning capacities of current VLMs instead of proposing a new model framework.
    We present a large-scale reasoning-based affordance segmentation benchmark, RAGNet, for general grasping. It is collected from diverse sources and carefully annotated with affordance mask and reasoning instructions.
    \item 
    %We build a large-scale affordance segmentation database from various sources. It contains  273k images, 180 categories, and 26k reasoning instructions.
    We introduce an affordance-based grasping baseline, AffordanceNet, which bridges the gap between VLM-based affordance prediction and real-robot general grasping.
    %which includes a VLM for accurate open-world affordance capture and a grasping network for graper pose prediction.
    \item 
    %Extensive experiments are conducted, including two zero-shot affordance segmentation tasks, a simulation task, and a real-robot task.
    We conduct extensive experiments, including zero-shot and out-of-domain affordance segmentation, real-robot grasping evaluation, which show great performance.
\end{itemize}

% With the recent advancements of multi-modal large language models (MLLM), the Embodied AI has been popular.
% Currently, the biggest problem of Embodied AI is the scarcity of annotation data.
% A mainstream solution is to employ Internet demonstration videos for action mining.
% However, this method lacks fine-grained perception, which will be a long-term goal.

% \begin{table*}[t]
% \centering
% \resizebox{0.5\linewidth}{!}{
% 		\setlength\tabcolsep{2pt}
% 		\renewcommand\arraystretch{1}
%     \begin{tabular}{l|cccc}
%     \hline \thickhline
%     Data source & Images & Domain & Annotation & Reasoning \\
%     \hline
%     HANDAL~\cite{handal} & 211k & Wild & \ding{182}  &\checkmark \\
%     Open-X~\cite{openx}  & 11.6k & Robot &\ding{184}\ding{185}\ding{186}   &  \\
%     GraspNet~\cite{graspnet} & 25k  & Robot &\ding{182}\ding{186}&\\
%     EgoObjects~\cite{egoobjects} & 12.7k & Ego & \ding{183}\ding{185}\ding{186}  & \checkmark \\
%     RLBench~\cite{rlbench} & 13k & Simulation & \noindent\ding{186}  &\\
%     \hline
%     \thickhline
%     \end{tabular}
%     }
%     \caption{\textbf{Training data and their annotation pipeline.} The details of  affordance mask annotation are in \S\ref{sec:affordance_annotation} and the generation details of reasoning instructions can be found in \S\ref{sec:reasoning_annotation}.}
%     \vspace{-5pt}
%     \label{tbl:source}
% \end{table*}

\section{Related Work}
\label{sec:related_work}

The investigation of affordances has a rich history in the computer vision and robotics communities~\cite{stark1994function,ferrari2007learning,kjellstrom2011visual,gibson2014ecological}.
For embodied agents to successfully interact with the functional components within a scene, they need the capability to comprehend visual affordances.
%Further, recognizing visual affordances necessitates an understanding of both the physical characteristics and the intended function of an object.
To advance research in this area, previous studies have dedicated significant resources to two main areas: benchmarks and algorithms.

%The problem of how to recognize partial affordance region has been studied in the past years due to its important application in robot manipulation.
%In this section, we will provide an overview of this field.

% \subsection{Affordance Prediction Benchmark}
\noindent\textbf{Benchmark.}
With the advances in deep learning, there has been a growing demand for the creation of large-scale affordance databases.
UMD~\cite{umd} stands out as a prior effort, which defines objects with effective affordances as those that an agent can grasp to produce an effect on another object.
In addition, it offers a dataset of 10k RGB-D images, each accompanied by a pixel-level affordance segmentation mask.
OPRA~\cite{fang2018demo2vec} proposes a different affordance learning pattern, which employs demonstration videos to guide the prediction of interaction region on a target image.
%It has 20,612 video clips.
It includes 20,612 video clips, each corresponding to an image and an annotation of interaction heatmap and action label.
Despite the pioneering efforts in affordance learning, the existing datasets~\cite{umd,nguyen2017object,luddecke2017learning,sawatzky2017weakly,chuang2018learning,fang2018demo2vec} still face constraints regarding affordance category diversity, image quality, and scene complexity.
Afterward, Luo \etal establish a large-scale affordance grounding dataset, AGD20k~\cite{agd20k} , which contains 20k exocentric images from 36 affordance categories.
Guo \etal build the HANDANL dataset~\cite{handal} for real-world robot manipulation, which provides precise handle annotation for 17 hardware and kitchen tool categories.
Recently, there are a variety of datasets on affordance prediction that have been introduced with different output formats, such as key points~\cite{3doi,li2024manipllm,ju2024robo,rashid2023language},  bounding box~\cite{vuong2023grasp,vuong2024language}, pixel-wise mask~\cite{li2024laso,li2024one,li2024learning}. 
However, most current research tends to focus on specific domains, such as robotics or first-person perspectives, which limits their applicability to other areas.
In contrast, our research aims to investigate the potential for open-world generalization in affordance prediction by leveraging extensive data.

%Affordance prediction has various output formats, such as keypoints,  bounding box, pixel-wise mask~\cite{luddecke2017learning,chuang2018learning,li2024laso,li2024one,li2024learning},  and Gauss map. 

% \subsection{Affordance Prediction Algorithm}
\noindent\textbf{Algorithm.}
In the deep learning era, the common methods use supervised learning for affordance prediction ~\cite{nguyen2017object,luddecke2017learning,do2018affordancenet,chuang2018learning,handal}.
However,  these approaches struggle to generalize to domains outside of their training environments.
To tackle this issue, certain studies utilize transfer learning~\cite{agd20k,li2023locate,yang2023grounding,luo2023leverage} or self-supervised learning techniques~\cite{chen2023affordance} to enable affordance prediction from data originating in different domains.
With the significant progress of VLM, attempting these foundation models in affordance prediction has attracted much interest in the research community~\cite{affordancellm,manipvqa,li2024one,li2024manipllm,yu2024uniaff,xu2024naturalvlm}.
However, these methods usually learn from limited affordance demonstration data, which opens up opportunities for further research into how large-scale data can impact generalization in open-world scenarios and enhance knowledge reasoning.

\section{Dataset}
\label{sec:dataset}

In this section, we will present how to build the large-scale reasoning-based affordance segmentation benchmark RAGNet.
%Toward this goal of exploring reasoning-based affordance data at scale, we first gather extensive data from various sources, as detailed in  \S\ref{sec:data_source}.
First, we gather extensive data from various sources, as detailed in  \S\ref{sec:data_source}.
Then, we introduce five tools to annotate this data with grasping-oriented affordance masks according to the original dataset characteristics in  \S\ref{sec:affordance_annotation}.
Meanwhile, we offer detailed instructions for reasoning-based affordance segmentation in  \S\ref{sec:reasoning_annotation}.
Lastly, we establish multiple validation benchmarks for testing grasping-oriented affordance segmentation in  \S\ref{sec:evaluation_metrics}.
The data annotation pipeline is illustrated in Fig.~\ref{fig:framework}.
%Finally, we deliver a comprehensive analysis of our dataset in  \S\ref{sec:dataset_statis}.

\begin{table}[t]
\centering
\resizebox{\linewidth}{!}{
		\setlength\tabcolsep{4pt}
		\renewcommand\arraystretch{1}
    \begin{tabular}{l|ccccc}
    \hline \thickhline
    Data source & Domain & Annotation & Rea. Inst. & Categories \\
    \hline
    HANDAL~\cite{handal}  & Wild & \ding{182}  & 8.5k & 17\\
    Open-X~\cite{openx}  & Robot &\ding{184}\ding{185}\ding{186}  & - & 124  \\
    GraspNet~\cite{graspnet}  & Robot &\ding{182}\ding{186} & - & 32\\
    EgoObjects~\cite{egoobjects}  & Ego & \ding{183}\ding{185}\ding{186}  & 17.4k & 74  \\
    RLBench~\cite{rlbench}  & Simulation & \noindent\ding{186}  & - &10\\
    \hline
    \thickhline
    \end{tabular}
    }
    \caption{\textbf{Details of training data annotation in RAGNet.} The  affordance mask annotation are in \S\ref{sec:affordance_annotation} and the generation details of reasoning instructions (Rea. Inst.) can be found in \S\ref{sec:reasoning_annotation}.}
    \vspace{-5pt}
    \label{tbl:source}
\end{table}

\subsection{Data Source}
\label{sec:data_source}

%Given that robotics can function across various embodiment realms, including real-world settings, robotic hardware, and even simulated environments, we collect a broad spectrum of data. 
Given that robotics can function across various embodiment realms, including real-world settings and robotic hardware, we collect a broad spectrum of data. 
The dataset comprises sources such as wild data (\ie, HANDAL~\cite{handal}), real robots (\ie, Open-X~\cite{openx}, GraspNet~\cite{graspnet}), and ego-centric data (\ie, EgoObjects~\cite{egoobjects}). 
For a fair comparison environment, we consider a collection of simulation data (\ie, RLbench~\cite{rlbench}).
We gather 273k images as shown in Table~\ref{tbl:source}.
%We retain 273k images, whose distribution is listed in Table~\ref{tbl:source}.
%We retain 300k images with pixel-wise mask annotation, whose distribution is listed in Figure.
%Besides, we also include several general datasets, like ADE20k~\cite{ade20k}, COCO-Stuff~\cite{cocostuff}, PACO~\cite{paco}, and \etc, for combined training.

\subsection{Affordance Map Annotation}
\label{sec:affordance_annotation}

%Before affordance annotation produced, we first class the classes into objects with handle and objects without handle.
To facilitate affordance segmentation with reduced manual input, we develop a suite of annotation tools.
The annotation workflow is designed to be adaptable, taking into account the functional and physical characteristics of various object categories.
For instance, when dealing with an object like a soda can, the robotic system typically grasps the entire object, necessitating full object annotation.
On the other hand, an object such as a wok requires precise annotation of its handle for effective grasping.
This tool suite comprises five annotation tools, each with its own level of priority.

\begin{figure}
    \centering
    \vspace{-4pt}
    \includegraphics[width=\linewidth]{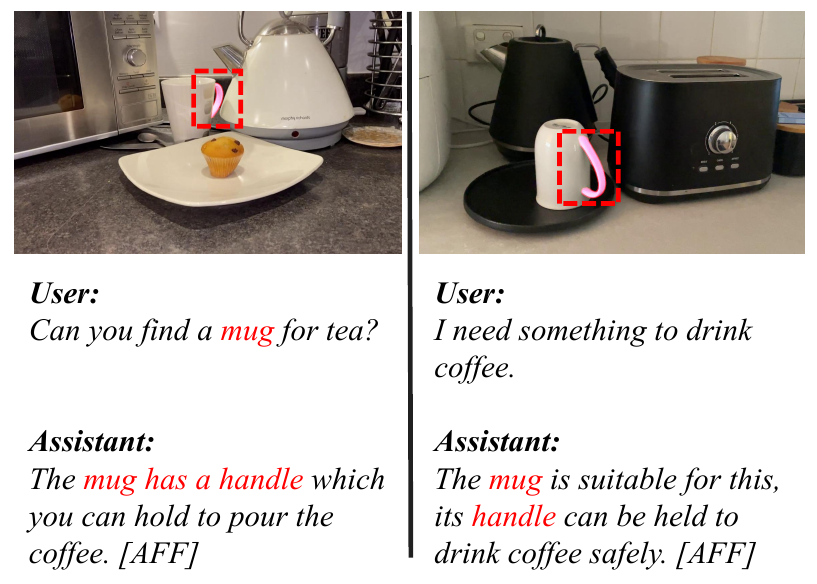}
    \vspace{-20pt}
    \caption{\textbf{Easy (left) \textit{v.s.} Hard (right) Reasoning Instruction.} The hard version has no category name itself.}
    \label{fig:reasoning_seg_gt}
    \vspace{-4pt}
\end{figure}

\noindent\ding{182} \textit{Original mask}:
Certain datasets specify affordance segmentation with precision (such as HANDAL dataset~\cite{handal}), while others, like those featuring objects without handles (\eg, computer mouse, soda can, and pen), do not require detailed affordance masks for grasping. In these cases, we utilize the original masks as our affordance annotations without further refinement.

\noindent\ding{183} \textit{SAM2}: 
For objects that lack handles, SAM2~\cite{sam2} can still be utilized to generate a mask, when only the ground-truth bounding box is available (like EgoObjects~\cite{egoobjects}).

\noindent\ding{184} \textit{Florence2 + SAM2}: 
Due to the presence of language instructions (like Open-X~\cite{openx}), Florence2~\cite{florence} which generates polygon boxes coupled with SAM2 can be used to produce an affordance map for objects that are handle-free.

\noindent\ding{185} \textit{VLPart + SAM2}: 
VLPart~\cite{vlpart} enables part-level recognition  (\ie, knife handle and mug handle), hence we leverage it and SAM2 for segmenting object handles if the corresponding category has been trained within VLPart.

\noindent\ding{186} \textit{Human (+ SAM2)}: 
If the above four tools fail to complete the affordance segmentation accurately, we will consider manual affordance annotation. The use of SAM2 is optional, particularly when dealing with video sequences.

According to the original annotation information offered by data sources, we employ different compositions for affordance mask annotation.
The composition details are listed in Table~\ref{tbl:source}.
%Taking the EgoObjects dataset as an example, its ground-truth bounding box can be directly used to generate masks with tool \ding{183}.
%More annotation details, including the tool arrangement difference in terms of subset and category, are included in the supplementary.
We incorporate more detailed annotations, encompassing variations in tool arrangement across subsets and categories, within the supplementary.
%Once the mask annotation is finished, we ask annotators to manu remove the inaccurate images.
Besides, we provide several representative examples in Fig.~\ref{fig:dataset} and more annotation examples can also be found in the supplementary.

\subsection{Reasoning Instruction Annotation}
\label{sec:reasoning_annotation}
As previously mentioned, the current VLM possesses compelling reasoning capabilities. 
To harness these capabilities for affordance reasoning, we construct a set of instructions.
In this section, we introduce three types of instructions, including one template-based and two reasoning-based.
%with representative examples provided in Table~\ref{tbl:instructions}.

The first kind of instruction is \textit{template-based}.
For instance, a template is ``Please segment the affordance map of \texttt{<category\_name>} in this image''.
This template can be utilized in our entire dataset for affordance prediction.
%More templates are included in the supplementary.
%Different from previous methods that employ templates, our generated instructions are diverse.
The second category consists of \textit{easy reasoning-based} instructions that are based on straightforward reasoning.
A key aspect of these instructions is the explicit mention of the object being referred to.
% The second one is \textit{easy reasoning-based} instructions. 
% Their main feature is including the referent object name itself.
In contrast to template-based approaches and previous studies~\cite{lisa,affordancellm,manipvqa},
the third category comprises \textit{hard reasoning-based} instructions, which do not include the category name.
% We construct two types of reasoning-based instructions, named easy mode and hard mode.
% The instructions in easy mode include the referent object name itself, while the hard mode does not contain the category name.
For precise identification of the target object, the hard mode instructions utilize a functional description.
Fig.~\ref{fig:reasoning_seg_gt} presents a typical example.
Towards grasping a mug, the easy instruction might be ``Can you find a mug for tea'', while the hard instruction is ``I need something to drink coffee''.
% For instance, an easy question might be "Please give me a hammer".
% In contrast, a hard question could be "We need to drive a nail into wood".

%To generate these instructions, we make full use of LLM.
% For generating these reasoning-based instructions with minimal human effort and computation sources, we take advantage of GPT-4.
% The used detailed prompt is shown in the supplementary.
% From our statistics, we create 211k easy reasoning-based instructions  and 211k hard reasoning-based for HANDAL.
% Moreover, we create 10k easy reasoning-based for EgoObjects.
To produce these reasoning-based instructions with minimal human labor and computational resources, we make full use of the capabilities of GPT-4~\cite{achiam2023gpt}. The specific prompt utilized is detailed in the supplementary material. 
%According to our data, we have crafted 211k easy logic-driven instructions and an equal number of hard ones for the HANDAL dataset. Additionally, we have developed 10k easy reasoning-based instructions for the EgoObjects dataset.
According to our data, we craft 8.5k hard instructions for the HANDAL dataset, 12.7k easy ones and 4.7k hard ones for the EgoObjects, totaling 26k reasoning-based instructions.

\begin{table}[t]
\centering
\resizebox{\linewidth}{!}{
		\setlength\tabcolsep{4pt}
		\renewcommand\arraystretch{1}
    \begin{tabular}{l|cccc}
    \hline \thickhline
    Validation Set & Images & Zero-shot  & Reason.  & Anno. \\
    \hline
    HANDAL~\cite{handal} & 65k & - & N/A & \ding{182} \\
    HANDAL$^{\dagger}$~\cite{handal} & 1k & -  &  N/A & \ding{182}  \\
    GraspNet \texttt{seen}~\cite{graspnet} & 1k & - &  N/A   &\ding{182}\ding{186} \\
    \rowcolor{gray!30} % 设置第一行的背景色为灰色
    GraspNet \texttt{novel}~\cite{graspnet} & 1k  & \checkmark &  N/A  &\ding{182}\ding{186}\\
        \rowcolor{gray!30} % 设置第一行的背景色为灰色
    3DOI~\cite{3doi} & 1k & \checkmark  &  N/A &\ding{186}\\
    \hline
    HANDAL$^{\dagger}$~\cite{handal} & 1k  & -  & easy & \ding{182} \\
    HANDAL$^{\dagger}$~\cite{handal} & 1k  & -  & hard & \ding{182} \\
        \rowcolor{gray!30} % 设置第一行的背景色为灰色
    3DOI~\cite{3doi}  & 1k  &  \checkmark& easy  &\ding{186}  \\
    \hline
    \thickhline
    \end{tabular}
    }
    \caption{\textbf{Details of validation set.} $^\dagger$ means a HANDAL subset. \colorbox{gray!20}{Gray}  means zero-shot setting. 
    As highly replicated images, we randomly select 1k images from the source dataset for validation.
    %1k images are randomly selected as replicated 
    }
    \vspace{-5pt}
    \label{tbl:validation}
\end{table}

\subsection{Evaluation Dataset}
\label{sec:evaluation_metrics}

% To evaluate the open-world ability of the affordance prediction model, the zero-shot evaluation is very important.
% Here, we design two kinds of zero-shot evaluation validation.
% The first one is to evaluate the ability of an affordance model to generalize to unseen object categories.
% Here we select the \texttt{novel} split of the GraspNet~\cite{graspnet}.
% The second one is to evaluate the ability of an affordance model to generalize to unseen data domains.
% Here, we choose 3OI~\cite{3doi}, which sources from Articulation~\cite{qian2022understanding},  EpicKitchen~\cite{damen2022rescaling} and Taskonomy~\cite{zamir2018taskonomy}, ensuring that here is no overlap between training the validation set.
To assess the open-world generalization of our affordance prediction model, 
%zero-shot evaluation plays a crucial role.
%In this work, 
we contribute two distinct zero-shot affordance evaluation scenarios.
The first scenario evaluates the affordance model by extending its predictions to object categories not encountered during training. 
%For this, we utilize the \texttt{novel} subset of the GraspNet dataset~\cite{graspnet}.
The second scenario evaluates the model generalization ability across different data domains. For this purpose, we select the 3DOI dataset~\cite{3doi}, which includes data from Articulation~\cite{qian2022understanding}, EpicKitchen~\cite{damen2022rescaling}, and Taskonomy~\cite{zamir2018taskonomy}, ensuring that there is no overlap with the training data in the validation set.

%In summary,  we present four validation sets for validating affordance segmentation, as shown in Table~\ref{tbl:validation}.
We present four different validation sets in Table~\ref{tbl:validation}.
Due to high-similarity images, we construct a subset from the HANDAL, which is marked with `$^\dagger$'.
For the same reason, other validation sets also utilize 1k images from the source dataset.
HANDAL and GraspNet \texttt{seen} represent the object categories and image domains that the model has been trained on.
GraspNet \texttt{novel} represents the unseen object category, while 3DOI refers to the unseen image domain.
%This is because  HANDAL test set has lots of replicated images and small subset can speed up evaluation process.
In addition, we provide three reasoning-based affordance segmentation validation sets, as illustrated in Table~\ref{tbl:validation}.
The reasoning instruction version used in each subset is marked as ``easy'' or ``hard''.
We utilize the generalized Intersection over Union (gIoU) and complete Intersection over Union (cIoU) as our primary metrics.
% \subsection{Dataset Statistics}
% \label{sec:dataset_statis}

\begin{table*}[t]
\centering
\resizebox{\linewidth}{!}{
		\setlength\tabcolsep{6pt}
		\renewcommand\arraystretch{1}
\begin{tabular}{l|cc|cc|cc|cc|cc}
    \toprule
    \multicolumn{1}{l|}{\multirow{2}*{Method}} & \multicolumn{2}{c|}{HANDAL} & \multicolumn{2}{c|}{HANDAL$^\dagger$ } & \multicolumn{2}{c|}{GraspNet \texttt{seen}} & \multicolumn{2}{c|}{\cellcolor{gray!30} GraspNet \texttt{novel}} & \multicolumn{2}{c}{\cellcolor{gray!30}3DOI} \\
    & gIoU & cIoU & gIoU & cIoU & gIoU & cIoU & \cellcolor{gray!30}gIoU & \cellcolor{gray!30}cIoU & \cellcolor{gray!30}gIoU & \cellcolor{gray!30}cIoU  \\
    \midrule
     \multicolumn{11}{c}{\textit{Foundation Model without LLMs}} \\
    \midrule
    % VLPart~\cite{vlpart} + SAM2~\cite{sam2} & 40.9 & 28.9 & 40.7& 27.6 & 55.5 & 42.7 & \textbf{55.3} & \textbf{49.0} & \textbf{52.4} & \textbf{51.4}  \\
        VLPart~\cite{vlpart} + SAM2~\cite{sam2} & 40.9 & 28.9 & 40.7& 27.6 & - & - & - & - & - & - \\
    % Grounding DINO~\cite{gdino} + SAM2~\cite{sam2} & 34.7 & 26.8 & 34.9 & 26.9 & 40.8 & 31.4 & 45.7  & 42.7 & 41.5 & 44.8 \\
        Grounding DINO~\cite{gdino} + SAM2~\cite{sam2} & 34.7 & 26.8 & 34.9 & 26.9 & - & - & - & - & - & -  \\
    % Florence 2~\cite{florence} + SAM2~\cite{sam2} & 39.7 & 22.4 & 39.4 & 22.5 & 27.5 & 19.4 & 22.6 & 15.3 & 40.4 & 25.4 \\
        Florence 2~\cite{florence} + SAM2~\cite{sam2} & 39.7 & 22.4 & 39.4 & 22.5& - & - & - & - & - & -  \\
    \midrule
    \multicolumn{11}{c}{\textit{Generalist MLLMs}} \\
    \midrule
    LISA~\cite{lisa}  & 16.2 & 12.0 & 15.4 & 11.8 & 17.7 & 17.7 & 25.2 & 24.1 & 21.5 & 13.7  \\
    GLaMM~\cite{glamm} & 24.9 & 17.2& 25.1& 17.0& 21.6 & 10.5 & 19.2 & 8.6 & 19.7 & 14.1  \\
    % \midrule
    % \multicolumn{9}{c}{\textit{Specialist MLLMs}} \\
    % \midrule
    % ManipVQA~\cite{manipvqa} & 17.8 & \\
    \rowcolor{cyan!10} AffordanceNet (Ours) & \textbf{60.3} & \textbf{60.8} & \textbf{60.5} & \textbf{60.3}& \textbf{63.3} &\textbf{64.0} & \textbf{45.6} &\textbf{33.2} & \textbf{37.4} &\textbf{37.4}\\
    \bottomrule
    \end{tabular}
    }
    \caption{\textbf{Quantitative results on affordance segmentation.} 
    %$^\dagger$ means a small-scale subset (1k images) from HANDAL test (65k images). 
    %$*$ represents that the model architectures are the same with LISA. 
    We use the fixed format of ``\texttt{<categoty\_name>} handle'' in foundation models without LLMs, while utilize ``affordance map of \texttt{<categoty\_name>}'' in generalist MLLMs.
    %Even though our model cannot outperform foundation models over some datasets, it contains a complex reasoning ability (see Table~\ref{tbl:reasoning_affordance}).
    % The best results are in bold.
    \colorbox{gray!20}{Gray}  means zero-shot benchmark. 
    }
    \vspace{-8pt}
    \label{tbl:affordance_segmentation}
\end{table*}

\begin{table}[t]
\centering
\resizebox{\linewidth}{!}{
		\setlength\tabcolsep{4pt}
		\renewcommand\arraystretch{1}
\begin{tabular}{l|cc|cc|cc}
    \toprule
    \multicolumn{1}{l|}{\multirow{2}*{Method}} & \multicolumn{2}{c|}{HANDAL$^\dagger$}  & \multicolumn{2}{c|}{\cellcolor{gray!30}Grasp. \texttt{novel}} & \multicolumn{2}{c}{\cellcolor{gray!30}3DOI}\\
    & gIoU & cIoU & \cellcolor{gray!30}gIoU & \cellcolor{gray!30}cIoU & \cellcolor{gray!30}gIoU & \cellcolor{gray!30}cIoU  \\
    \midrule
    % \multicolumn{7}{c}{Foundation Model without LLMs} \\
    % \midrule
    % Grounding DINO + SAM2 \\
    % Florence 2 + SAM 2 \\
    % \midrule
    % \multicolumn{7}{c}{Generalist MLLMs} \\
    % \midrule
    LISA & 16.2 & 12.0 &25.2 & 24.1 &   21.5 &   13.7 \\
    + HANDAL & 56.3 & 54.9 &  16.6 &   14.4 &   18.0 &   12.2 \\
    + Open-X  & 59.3 & 56.7 &  19.2 &  18.4 &   24.5 &   16.0 \\
    + EgoObejcts & 61.8 & 61.6 &   8.0 &  6.9 &  35.5 &   34.6\\
    + GraspNet & 61.7 & 61.7 &  51.5 &  38.5 &  40.9 &   41.8\\
    + Reasoning & 56.5 & 55.4 &  43.0&  33.8 &  36.8 &  40.2\\
    + RLBench & 56.7 & 55.0 &   42.8 &   33.2 &  36.5 &  39.9  \\
    % + RLBench \\
    % \midrule
    % + Prompt \\
    % + Token \\
     \rowcolor{cyan!10}  Ours & 60.5 & 60.3 &  45.6 &  33.2 &   37.4 &  37.4 \\
    % \midrule
    % \multicolumn{7}{c}{Specialist MLLMs} \\
    % \midrule
    % ManipVQA \\
    % Ours \\
    \bottomrule
\end{tabular}
}
    \caption{\textbf{Ablation study of data on affordance segmentation.} Each data is added one by one. Compared to ``+ RLBench'', our final model is enhanced by task-specific modifications, including specialized system prompt and unique  \texttt{<AFF>} token (see \S\ref{sec:affordancevlm}).}
    \vspace{-8pt}
    \label{tbl:ablation}
\end{table}

\section{AffordanceNet}

To achieve the goal of open-world grasping, we propose a comprehensive framework, named AffordanceNet.
%It includes two important components: VLM for open-word affordance perception and rule-based strategies for robotic manipulation.
This model consists of two key components: AffordanceVLM for predicting affordance segmentation mask and pose generation for transforming the mask into grasper position in 3D space.
The overall framework is illustrated in Fig.~\ref{fig:framework}.
% After the collection of large-scale reasoning-based affordance segmentation data, we base them to train a VLM-based affordance prediction model.

\subsection{AffordanceVLM}
\label{sec:affordancevlm}

% \noindent\textbf{Model Architecture.} Our AffordanceVLM is based on the widely-adopted vision-language segmentation model LISA~\cite{lisa}.
% First, we choose LLaVA-7B~\cite{llava} as our backbone model, comprising an image encoder based on ViT-CLIP~\cite{radford2021learning}, a linear layer projector, and a pre-trained Large Language Model (LLM). 
% The input image is encoded by the image encoder and then projected into the LLM's space with the help of the projector. Meanwhile, the text is tokenized by a text tokenizer. These two sets of features are concatenated and input into the LLM (\ie, Vicuna-7B~\cite{zheng2023judging}), which generates a response marked by a special token \texttt{<SEG>}.
% Then, we employ SAM~\cite{kirillov2023segment} as a mask decoder to transform the special token into a pixel-wise mask for affordance prediction.
% We training this model on our provided massive affordance data to build AffordanceVLM.

Our AffordanceVLM is based on the vision-language segmentation model LISA~\cite{lisa} and incorporates two essential task-specific modifications to enhance affordance prediction: (1) developing a specialized system prompt, and (2) introducing a unique \texttt{<AFF>} token.

Specifically, we first process the input image using an image encoder (\ie, ViT-CLIP~\cite{radford2021learning}), which is then projected into the LLM's embedding space via a linear layer projector. Meanwhile, the language prompt is tokenized by a text tokenizer, where each affordance instruction is enhanced by ``You are an embodied robot."
The resulting image and text features are concatenated and fed into the LLM (\ie, Vicuna-7B~\cite{zheng2023judging}). 
For general segmentation, the LLM generates a response that includes a special token \texttt{<SEG>}. 
However, since  \texttt{<SEG>} is a token within the LLM's vocabulary, its representation is confined to a fixed feature space. This limitation restricts its representation capacity, thereby affecting the quality of the decoded mask. To address this issue, we introduce another special token \texttt{<AFF>} to enrich the original mask embedding.
The underlying motivation is to explicitly direct the final mask embedding to focus more on affordance-specific language expressions.
Finally, SAM~\cite{kirillov2023segment} is used as a mask decoder to convert this mask embedding into a pixel-wise mask.

% \noindent\textbf{Overall Architecture.} 
% First, we choose LLaVA-7B~\cite{llava} as our backbone model, comprising an image encoder based on ViT-CLIP~\cite{radford2021learning}, a linear layer projector, and a pre-trained Large Language Model (LLM). 
% The input image is encoded by the image encoder and then projected into the LLM's space with the help of the projector. Meanwhile, the text is tokenized by a text tokenizer. These two sets of features are concatenated and input into the LLM (\ie, Vicuna-7B~\cite{zheng2023judging}), which generates a response marked by a special token \texttt{<AFF>}.
% Then, we employ SAM~\cite{kirillov2023segment} as a mask decoder to transform the special token into a pixel-wise mask for affordance prediction.
% We training this model on our provided massive affordance data to build AffordanceVLM.

\noindent\textbf{Implementation Details.}
% Even though the various subsets in our data contain varying numbers of images, we uniformly sample from each subset.
% Beyond our reasoning-based affordance segmentation data, we also incorporate a variety of generic segmentation datasets into our training. This diverse generic set includes data for semantic segmentation (\eg, ADE20k~\cite{ade20k}, COCO-Stuff~\cite{cocostuff}, PACO~\cite{paco}), referring segmentation (\eg, RefCOCO~\cite{refcoco}), and reasoning-based segmentation (\eg, ReasonSeg~\cite{lisa}), which use \texttt{<SEG>} token.
Beyond our reasoning-based affordance segmentation data, we also incorporate a variety of generic segmentation datasets into our training, which use \texttt{<SEG>} token.
During inference, we first extract the \texttt{<AFF>} token, followed by the \texttt{<SEG>} token.
More implementation details about data sampling and training settings can be found in the supplementary.
% We deploy eight NVIDIA A100 GPUs (80GB) to train our model, with a learning rate of  2e-5. 
% The training loss follows~\cite{lisa}, which uses binary focal loss for map prediction and cross-entropy loss for text output.
% We utilize a batch size of 40 without gradient accumulation.

\subsection{Pose Generator}
\label{sec:pose}

%After training the vision-language model using this large-scale affordance segmentation data, we observe that it exhibits enhanced zero-shot perception capabilities.
Once obtained AffordanceVLM that delivers precise affordance segmentation prediction, we start to apply the model to robotic grasping tasks, aiming to bridge the gap for the final steps in object grasping and manipulation. 
The impressive affordance segmentation results can be found in  \S\ref{sec:evaluation_affordance} and \S\ref{sec:evaluation_reasoning}.
%Specifically, the model is capable of delivering precise affordance segmentation predictions even for domain data that were not part of the training process (see experiment results in \S\ref{sec:evaluation_affordance} and \S\ref{sec:evaluation_reasoning}).
%This discovery motivates us to apply the model to robotic manipulation tasks, aiming to bridge the gap for the final steps in object grasping and manipulation. 
In the following part, we will discuss more details about our grasp pose generator and the process of converting 2D affordance masks into the 3D pose estimates required for robotic arms.

% As illustrated in Fig.~\ref{fig:framework}, to accurately capture 3D information, we use an additional depth camera along with the RGB camera, that is RGB-D camera.
% Let $\bm{D}$ define the depth image, and $\bm{M}$ define the affordance mask.
% We crop the affordance region $\bm{A}$ with a binary multiplication $\otimes$:  $\bm{A} \!=\! \bm{D} \otimes \bm{M}$, and calculate the 2D coordinates of these remained pixels, defined as $\bm{A}_{2D}$.

As illustrated in Fig.~\ref{fig:framework}, we project the depth maps into 3D space for precise grasp pose generation. Let $P$ denotes the set of points on the 2D image. Firstly, we filter the affordance region by applying the affordance mask $M$ to $P$ through a binary multiplication $\otimes$, resulting in $ \hat{P} = P \otimes M $. Subsequently, for each 2D position $(u, v)$ within the point set $ \hat{P}$, we have
\begin{flalign}
\small
\vspace{-30pt}
    &\left[\begin{matrix}x \\y \\z \\1\end{matrix}\right] = T \cdot K^{-1}  \left[\begin{matrix}u \times d \\v \times d \\d \\1\end{matrix}\right],
    \label{equ_2}
\vspace{-30pt}
\end{flalign}
where d is the along the axis orthogonal to the image plane, $(x, y, z)$ are the corresponding world coordinates. K $\in \mathbb{R}^{4\times4}$ and T $\in \mathbb{R}^{4\times4}$ represent the camera intrinsic and extrinsic parameters, respectively.  
%where $\otimes$ represents binary multiplication.
%Once getting the 3D object, we can use various methods to control the grasp.
After getting the 3D position of object affordance, we can use various grasping models to generate grasper position.
Finally, the grasper arrive the requested position and grasp the target object.

\begin{figure*}[t]
    \centering
    \includegraphics[width=\linewidth]{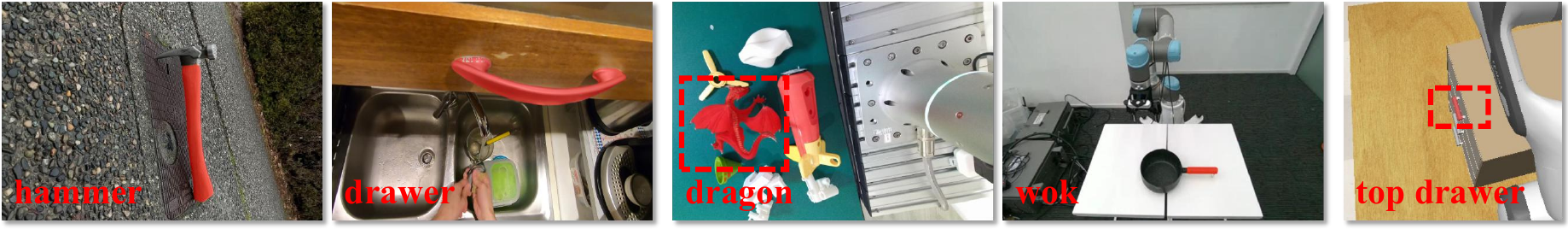}
    \vspace{-22pt}
    \caption{\textbf{Affordance segmentation from our AffordanceNet}. Even though they source from various data sources, such as wild, robot, ego-centric and simulation, our model can accurately capture their affordance region. More visualizations are included in supplementary.}
    \label{fig:results_aff_seg}
        \vspace{-4pt}
\end{figure*}

\begin{figure*}[t]
    \centering
    \includegraphics[width=\linewidth]{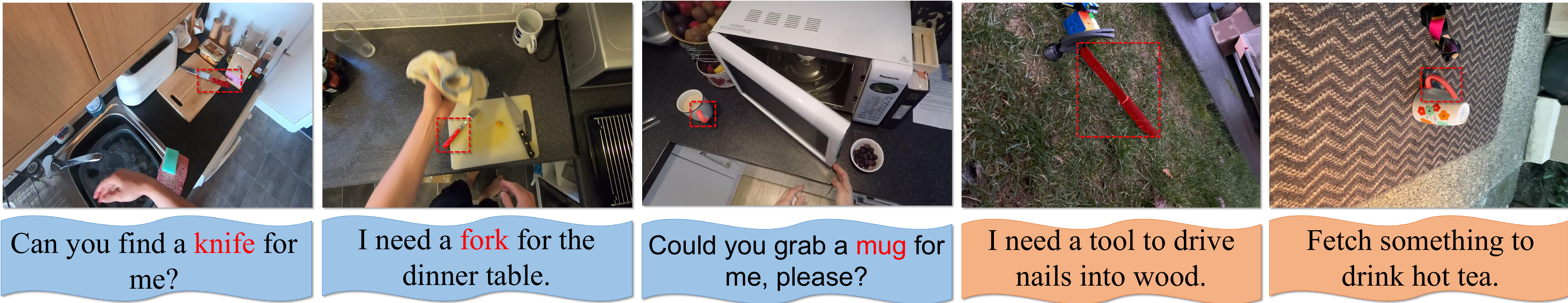}
        \vspace{-22pt}
    \caption{\textbf{Reasoning-based affordance segmentation from our AffordanceNet}.  The left examples represent easy reasoning-based instructions with referent name, while the right are hard instructions that include object function or intention rather than the name itself.}
    \label{fig:results_reason_seg}
            \vspace{-6pt}
\end{figure*}

\begin{table}[t]
\centering
\resizebox{\linewidth}{!}{
		\setlength\tabcolsep{4pt}
		\renewcommand\arraystretch{1}
\begin{tabular}{l|cc|cc|cc}
    \toprule
    \multicolumn{1}{l|}{\multirow{2}*{Method}} & \multicolumn{2}{c|}{HANDAL (easy)}  & \multicolumn{2}{c|}{HANDAL (hard)} & \multicolumn{2}{c}{\cellcolor{gray!30}3DOI}\\
    & gIoU & cIoU & gIoU & cIoU & \cellcolor{gray!30}gIoU & \cellcolor{gray!30}cIoU  \\
    \midrule
    G-DINO & 3.6 & 3.0 & 3.4 & 3.1 & 4.1 & 3.9 \\
    LISA & 15.5 & 11.9 & 12.3 & 8.1 &12.3 &  8.1  \\
    GLaMM & 4.7 & 3.5 & 5.0 & 3.5 & 4.4 &2.9\\
    \rowcolor{cyan!10}  Ours & \textbf{58.3} & \textbf{58.1}  & \textbf{58.2} & \textbf{57.8} &\textbf{38.1} &\textbf{39.4} \\
    \bottomrule
\end{tabular}
}
    \caption{\textbf{Quantitative results on reasoning-based affordance segmentation.}  `G-DINO' is the short name of Grounding-DINO. They all use reasoning-based instructions as language prompts.}
    \vspace{-5pt}
    \label{tbl:reasoning_affordance}
\end{table}

\begin{figure*}[t]
        \centering
    \includegraphics[width=0.95\linewidth]{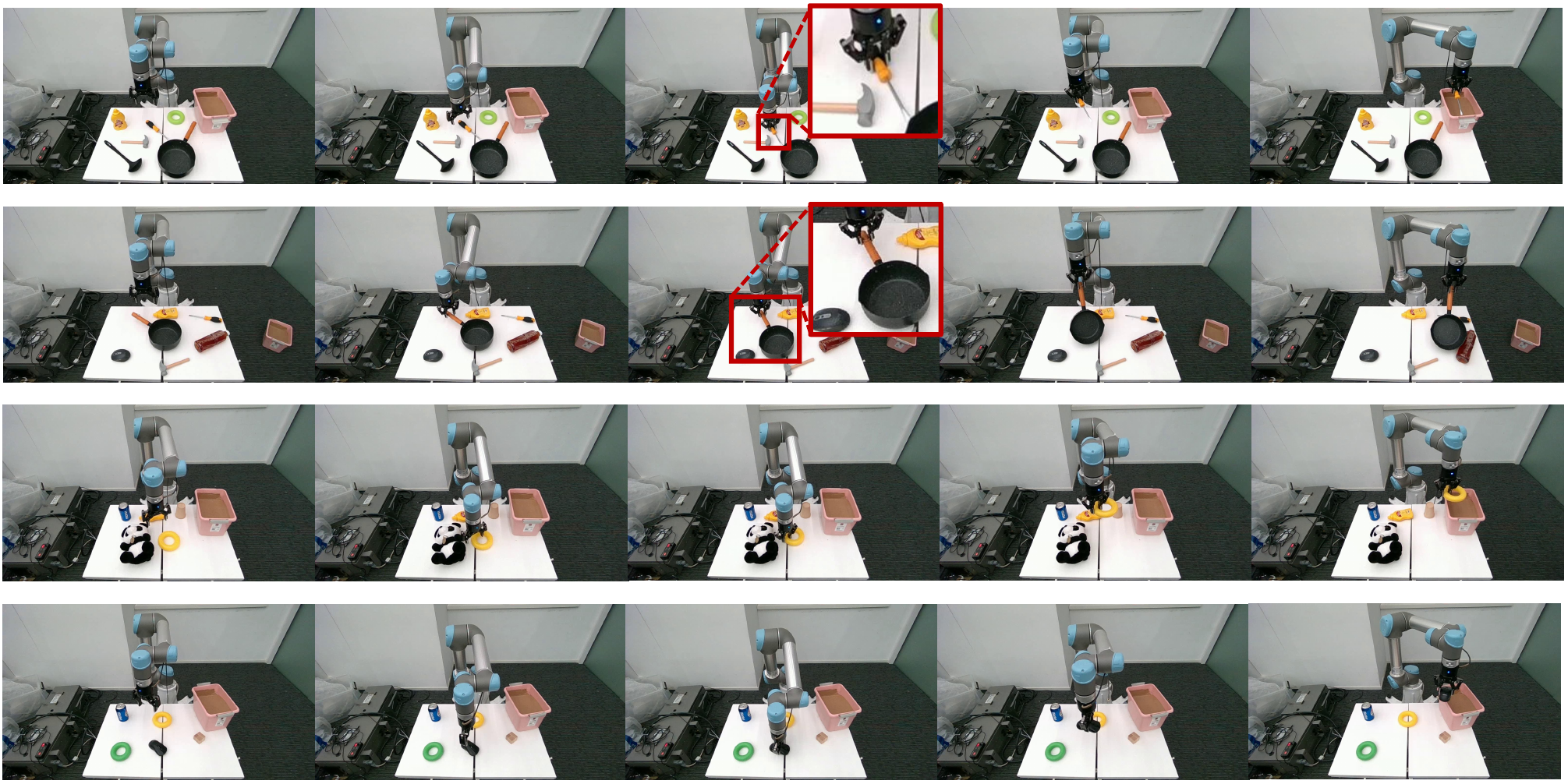}
        \vspace{-10pt}
    \caption{\textbf{Object grasping results from our AffordanceNet on robot arm UR5}. The instructions are ``\textit{I need a screwdriver for repairing}'', ``\textit{Can you hand me the wok, please?}'', ``\textit{Give me the circle}'', ``\textit{Please hand me a computer mouse}'', respectively.}
    \vspace{-4pt}
    \label{fig:vis_robot}
\end{figure*}

\begin{table*}[t]
\centering
\resizebox{\linewidth}{!}{
		\setlength\tabcolsep{6pt}
		\renewcommand\arraystretch{1}
\begin{tabular}{l|cccccccccc|c}
    \toprule
    % \multicolumn{1}{l|}{\multirow{2}*{Method}} & \multicolumn{2}{c|}{HANDAL}  & \multicolumn{2}{c|}{Grasp. novel} & \multicolumn{2}{c}{3DOI}\\
    % & gIoU & cIoU & gIoU & cIoU & gIoU & cIoU  \\
    Method & Can & Pen & Screwdriver & Hammer & Wok & Mouse & Circle & Toy & Spatula & Scissors & Average \\
    \midrule
GraspNet~\cite{graspnet} & 40\% & 10\% & 10\% & 20\% & 60\% & 50\% & 10\% & 60\% & 30\% & 30\% & 32\% \\
 \rowcolor{cyan!10}  AffordanceNet (Ours)  & \textbf{80\%} & \textbf{60\%} & \textbf{60\%} & \textbf{80\%} & \textbf{70\%} & \textbf{80\%} & \textbf{40\%} &\textbf{90\%} & \textbf{70\%} & \textbf{70\%} & \textbf{70\%}\\
    \bottomrule
\end{tabular}}
    \caption{\textbf{Average success rates on robotic grasping.} GraspNet cannot support language (see \S\ref{sec:real_robot}).  Each task is conducted by 10 times.}
    \vspace{-8pt}
    \label{tbl:results_robot}
\end{table*}

% \begin{figure*}[t]
%         \centering
%     \includegraphics[width=0.9\linewidth]{figs/results_simulation.pdf}
%         \vspace{-10pt}
%     \caption{\textbf{Object grasping results from our AffordanceNet on RLBench}. The instruction of the top video is ``\textit{Open the top drawer}'',  and the bottom one refers to ``\textit{Close the green jar}''.}
%     \label{fig:vis_simulation}
%         \vspace{-7pt}
% \end{figure*}

\section{Experiments on Visual Affordance}

To minimize unnecessary source expenditure, we initially validate the quality of affordance segmentation (\S\ref{sec:affordancevlm}) prior to object grasping (\S\ref{sec:pose}).

\subsection{Evaluation on Affordance Segmentation}
\label{sec:evaluation_affordance}

\noindent\textbf{Implementation Details.}
To evaluate the task of affordance segmentation, we implement various advanced open-sourced approaches with potential affordance segmentation ability.
In specific, we choose the foundation models without LLMs (\eg, VLPart~\cite{vlpart}, Grounding DINO~\cite{gdino}, Florence2~\cite{florence}), and the generalist MLLMs (\eg,  LISA~\cite{lisa}, GLaMM~\cite{glamm}). 
Since these foundation models only output bounding boxes or polygon boxes, we additionally employ SAM2~\cite{sam2} for mask refinement.
%Because the fuzzy concept of ``affordance map of an object'' is challenging to the foundation models without LLMs, we turn to ``object handle'' and evaluate them on only HANDAL dataset.
We load their official checkpoints for direct evaluation.
Given the ambiguous nature of the ``affordance map of \texttt{<category\_name>}" concept, which poses a challenge to the foundation models without LLMs, we shift to a fixed format of instruction ``\texttt{<category\_name>} handle" and evaluate them on only HANDAL dataset as each object have a handle.
In contrast, we employ ``affordance map of \texttt{<category\_name>}" when testing the generalist MLLMs.

\begin{table}[t]\small{
		% \vspace{-4pt}
		\centering
		\small
		\resizebox{0.48\textwidth}{!}{
			\setlength\tabcolsep{4pt}
			\renewcommand\arraystretch{1.0}
			\begin{tabular}{l|ccccc|c}
                \toprule
				% 			&\multicolumn{5}{c}{Frames}\\
				Method & Can & Pen & Screw. & Ham. & Wok & Average \\
				\midrule
                    VLPart~\cite{vlpart} & 70\% & 30\% & 40\% & 30\% & 0 & 34\%  \\ 
                    LISA~\cite{lisa} & 80\% & 40\% & 0 & 10\% & 0 & 26\%  \\
                    \midrule
                    Easy Reasoning & 70\% & 50\% & 50\% & 80\% & 60\% & 62\% \\
                    Hard Reasoning & 60\% & 40\% & 40\% & 60\% & 40\% & 48\% \\
                    \midrule
                    \rowcolor{cyan!10}   AffordanceNet & 80\%& 60\% & 60\% & 80\% & 70\% & 70\% \\
                    \bottomrule
		\end{tabular} }
		% \vspace{-8pt}	
            % \captionsetup{font={footnotesize}}
		\caption{\textbf{Ablation studies on real-robot grasping} in terms of different affordance prediction models (\ie, VLPart and LISA) and different instructions (\ie, easy reasoning and hard reasoning). }
		\vspace{-10pt}
		\label{table:real_robot_ablation}}
\end{table}

\noindent\textbf{Experiment Results.}
The main results on affordance segmentation are shown in Table~\ref{tbl:affordance_segmentation}.
%We can see that our model outperforms all other methods.
% We can see our model outperforms all other generalist MLLMs across all datasets, while achieving comparable scores compared to some foundation models over some datasets.
We can see our model outperforms all other competitors across all datasets.
Besides, the scores on the entire HANDAL dataset and its small subset are similar, guaranteeing the diversity and representative of the small subset.
Several visualizations across various domains are shown in  Fig.~\ref{fig:results_aff_seg}.
Here, the wok handle is accurately segmented even if its scene has never been encountered,  indicating remarkable open-world generalization.
%Besides, we provide several visualization results in Fig.~\ref{fig:results_aff_seg} across different data domains, such as wild data, robot data, ego-centric data, and simulation.

\noindent\textbf{Ablation Study.}
It is of interest to examine the impact of individual datasets on affordance segmentation tasks.
In Table~\ref{tbl:ablation}, we incrementally incorporate each dataset into the training process.
We can see that the absence of HANDAL data significantly impairs the model performance on the HANDAL test set.
%The same phenomenon can be observed in GraspNet.
Furthermore, the incorporation of reasoning data leads to a slight decline in performance metrics.
However, the introduction of task-specific modifications, such as a specialized system prompt and a unique \texttt{<AFF>} token, enhances model performance. 
\textit{Overall, our model shows powerful open-world affordance segmentation.}
%Besides, when adding reasoning data, the performance scores across all affordance segmentation are decreasing.

\subsection{Evaluation on Reasoning Affordance}
\label{sec:evaluation_reasoning}
%We also evaluate the generalist and specialist MLLMs on our reasoning-based affordance segmentation.
We also test the foundation models and generalist MLLMs on reasoning-based affordance segmentation datasets.
All the experiment settings are aligned with the above section.
%Since the foundation models without LLMs cannot support reasoning-based instructions, we do not provide experiment results in this part.

\noindent\textbf{Experiment Results.}
Despite this challenging task, the results in Table~\ref{tbl:reasoning_affordance} show our model outperforms other methods by a large margin.
We provide several reasoning-based qualitative results in Fig.~\ref{fig:results_reason_seg}.
The last two examples show that our model can predict precise hammer and mug handles even if there is no target mentioned in the instructions.
These results confirm the reasoning ability of our model.

\noindent\textbf{Remark.}
The above two vision experiments demonstrate that only our model excels in both affordance perception and reasoning in zero-shot domain. This provides a solid foundation for the subsequent object grasping tasks.
%Before object grasping, the two vision experiments indicate that our model shows powerful affordance perception and reasoning ability.

\section{Experiments on Object Grasping}

\subsection{Evaluation on Real Robot}
\label{sec:real_robot}

\noindent\textbf{Implementation Details.}
%To evaluate the open-world generalization, we design a set of experiments on real robot.
To evaluate the effectiveness of our model for open-world generalization in real-world environments, we introduce a series of manipulation experiments.
Specifically, we deploy UR5 robot arm with a third-person RGB-D camera (Intel RealSense).
We design 10 distinct grasping tasks, including grasping the can, pen, screwdriver, hammer, wok, mouse, circle, toy, spatula, scissors.
Half of them require accurately localizing the affordance region, like the screwdriver handle.
Each task is performed 10 times, and we report the average success rate.
%This is a zero-shot evaluation.
\textit{Note that, we never provide any demonstration images or videos from this scene for our model training.}
The well-trained AffordanceVLM is directly used for the zero-shot evaluation.
As for the pose generation, we follow GraspNet~\cite{graspnet} condition 3D affordance point cloud to generate a 3D grasp proposal for grasper operation.
%In this work, we employ UR5 as our hardware and task it three grasping instructions: ``Please grasp the screwdriver'', ``Please grasp the wok'', and ``Please grasp the computer mouse''.
We compare our model with a popular grasping model, GraspNet~\cite{graspnet}.
Since GraspNet lacks the capability for language-conditioned grasping, \textit{we ensure that only the target object remains on the table.}

\noindent\textbf{Experiment Results.}
%The success rates are listed in Table~\ref{tbl:results_robot}.
Table~\ref{tbl:results_robot} displays the performance comparison between ours and GraspNet.
Clearly, our model AffordanceNet has superior success rates, even in challenging and complicated environments.
%indicating the advantage of affordance segmentation.
%We can find that our AffordanceNet achieves higher success rates than GraspNet, and owns powerful open-world generalization ability.
Fig.~\ref{fig:vis_robot} highlights four sequences, which show accurate affordance perception and effective object grasping (\ie, screwdriver, wok, circle, and mouse). 
More grasping examples are in our supplementary.
%More grasping videos on novel object categories  (\eg, panda, toy, and circle) and comparisons with GraspNet can be found in our supplementary.
%More videos (\eg, panda, toy) and comparisons can be found in our supplementary.
%Some video sequences about grasping objects from our model are 
%We compare our AffordanceNet with GraspNet.

\noindent\textbf{Ablation Study.}
Table~\ref{table:real_robot_ablation} shows ablations of AffordanceNet on affordance prediction models (replaced by VLPart and LISA)  and instruction types (replaced by easy and hard reasoning-based instructions) using five tasks.
As demonstrated, AffordanceNet delivers superior affordance prediction while maintaining its reasoning capabilities.

\subsection{Evaluation on Simulation}
\noindent\textbf{Implementation Details.}
We conduct simulations based on RLBench sub-task~\cite{rlbench} to validate our approach,
%The obtained affordance mask is converted into actions for the robotic arm. 
including open drawer, close jar, and slide block to target. 
In practice, we divide each task into several keyframes, as follows: open drawer (3 keyframes), close jar (4 keyframes), slide block to target (5 keyframes). 
More details are in the supplementary.
\begin{table}[t]
\centering
\resizebox{\linewidth}{!}{
		\setlength\tabcolsep{8pt}
		\renewcommand\arraystretch{1}
\begin{tabular}{l|cc}
    \toprule
    % \multicolumn{1}{l|}{\multirow{2}*{Method}} & \multicolumn{2}{c|}{HANDAL}  & \multicolumn{2}{c|}{Grasp. novel} & \multicolumn{2}{c}{3DOI}\\
    % & gIoU & cIoU & gIoU & cIoU & gIoU & cIoU  \\
    Task & LLARVA~\cite{niu2024llarva} & \cellcolor{cyan!10}   AffordanceNet (Ours)\\
    \midrule
    Open drawer & 60\% & \cellcolor{cyan!10}  56\%\\
    Slide block to target & 100\% &  \cellcolor{cyan!10}  64\%\\
    Close jar & 28\% &  \cellcolor{cyan!10}  44\% \\  
    \textcolor{gray}{Average} & 62\% &  \cellcolor{cyan!10}  54.7\% \\
    \bottomrule
\end{tabular}
}

    \caption{\textbf{Success rates of our model on RLBench simulation.} 
    Each task is conducted by 25 episodes.
    }
    \vspace{-10pt}
    \label{tbl:results_simulation}
\end{table}

\noindent\textbf{Experiment Results.}
We conduct 25 episodes for each task and calculate its average success rate as the primary evaluation metric. 
The quantitative results are in Table \ref{tbl:results_simulation}.
%Visualization examples are included in supplementary.
Compared to another LLM-based method LLARVA~\cite{niu2024llarva} that is fine-tuned for a specific environment, our model focuses on stronger generalization. \textit{Therefore, achieving comparable performance on RLBench is quite satisfying.}

\section{Conclusion}

In this work, we presented a new large-scale and diverse reasoning-based affordance segmentation dataset, named RAGNet, which aims to advance the capabilities of general robotic grasping systems in varied open-world scenarios. 
%Our benchmark, comprising 273k images across 180 categories with 26k reasoning instructions, represents a significant step forward in providing the necessary data for training and evaluating affordance segmentation models that can generalize well to unseen objects and domains.
Further, we proposed the model AffordanceNet, a comprehensive framework including the AffordanceVLM and the grasping module, which uses our extensive affordance data to achieve open-world affordance capture and 3D grasper pose prediction, respectively. 
%In addition, we presented AffordanceNet, a holistic framework consisting of AffordanceVLM and the grasping module, which leverages our extensive affordance data to achieve accurate open-world affordance capture and 3D grasper pose prediction. 
Through extensive experiments, we observed the superior performance and generalization ability of our AffordanceNet in zero-shot affordance segmentation, reasoning-based affordance segmentation, real-robot grasping evaluation, and simulation tasks.
%showcasing its powerful open-world generalization ability.
\clearpage
{
    \small
    \bibliographystyle{ieeenat_fullname}
    \bibliography{main}
}
\clearpage
\setcounter{page}{1}
\maketitlesupplementary

\section{Details of Data  Annotation}
%As the original data sources provide different annotation information, such as masks from HANDAL~\cite{handal} and boxes from EgoObjects~\cite{egoobjects}, we consider different annotation compositions for our tools.
As the original data sources, such as HANDAL~\cite{handal}, Open-X~\cite{openx}, EgoObjects~\cite{egoobjects}, GraspNet~\cite{graspnet}, provide original annotation information (\eg, ground-truth boxes or masks), we make full use of them for minimal human intervention.
%and propose different annotation tool compositions.
%Besides, we focus on grasping-oriented objects, which consist of objects with handles and handle-free objects.
From these data, we emphasize grasping-oriented objects, encompassing both those with handles and those without.
Therefore, as described before, we design five annotation tools: \ding{182}: Original mask, \ding{183}: SAM2~\cite{sam2}, \ding{184}: Florence2~\cite{florence} + SAM2, \ding{185}: VLPart~\cite{vlpart} + SAM2, \ding{186}: Human (+ SAM2).
The details of how to compose these tools for one specific dataset are shown in Table~\ref{tbl:supp_train_data}.
In addition, Table~\ref{tbl:supp_train_data} provides the reasoning instruction annotation details.
\textit{In summary, our dataset RAGNet includes a broad range of data domains, categories, and reasoning instructions, establishing a robust basis for open-world grasping applications.}
%The tool composition details and category splitting are shown in Table~\ref{tbl:supp_train_data}.

\section{Affordance Annotation Examples}

\begin{figure*}
    \centering
    \includegraphics[width=\linewidth]{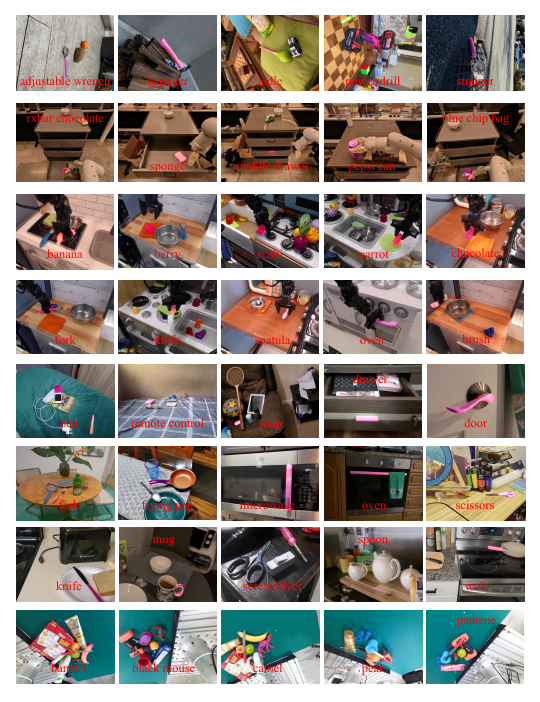}
    \vspace{-32pt}
    \caption{\textbf{More affordance segmentation examples from our RAGNet.} It covers various data sources, like robot, wild, and ego-centric domains.}
    \label{fig:supp_annotation}
\end{figure*}

%In this section, we present mode annotation examples.
%As seen, 
Since our benchmark RAGNet includes a significant number of grasping-oriented objects from various domains (like robot, wild, and ego-centric domains), we highlight this aspect by showcasing additional examples of affordance segmentation annotations in Fig.~\ref{fig:supp_annotation}.
%The candidate objects are first recognized if the handle is included, and then carefully annotated its affordance map.
For each affordance map annotation, the candidate objects are initially identified to determine if they possess a handle. 
Then, their affordance maps are carefully annotated according to our tool priority.
For example, the banana from Open-X~\cite{openx} is segmented using the combination of Florence2~\cite{florence} and SAM2~\cite{sam2} according to its original grasping instruction.
The knife handle from EgoObjects~\cite{egoobjects} can be accurately grounded using VLPart~\cite{vlpart}, and its output box can be further transformed into a pixel-wise mask using SAM2~\cite{sam2}.
%As for the handle of microwave from EgoObjects~\cite{egoobjects}, it is manually annotated due to the unavailable of suitable annotation tool.
Regarding the microwave handle in the EgoObjects dataset~\cite{egoobjects}, it has been annotated manually because there is no suitable tool available for automated annotation.
In conclusion, we collect a total of 273k diverse images along with their corresponding affordance annotations.

%In summary, our dataset provides diverse data domains and object categories for affordance segmentation prediction.
%For example, despite the robot arm occluding the banana in the third row, the corresponding affordance map is accurately annotated.

\section{Reasoning-based Affordance Examples}

More reasoning-based affordance segmentation examples are shown in Fig.~\ref{fig:supp_results}.
It contains two types of instructions, easy instructions and hard instructions.
As seen, the easy instructions include the target object name, while the hard ones only include functional descriptions without the object name.
These instructions are generated by GPT-4 and the corresponding prompts used in GPT-4 are listed in Table~\ref{table:supp_easy_reasoning} and Table~\ref{table:supp_hard_reasoning}.
The highlighted ``\textcolor{red}{words}'' are category names at most times.
We sometimes provide additional keywords about potential grasping action for some categories, for aligning the instructions with the image content.
For example, if the microwave is closed, we would assign the keywords ``\textcolor{red}{microwave, open the door}''. 
%To align the instructions with the image content, we provide additional keywords about motion for some categories

\begin{figure*}[h]
        \centering
    \includegraphics[width=\linewidth]{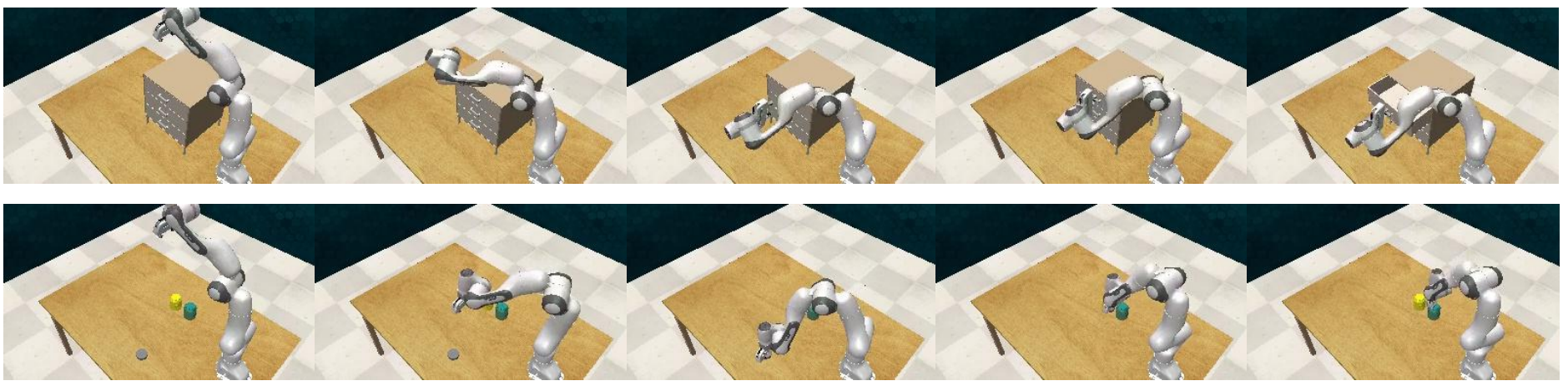}
        \vspace{-16pt}
    \caption{\textbf{Object grasping results from our AffordanceNet on RLBench}. The instruction of the top video is ``\textit{Open the top drawer}'',  and the bottom one refers to ``\textit{Close the green jar}''.}
    \label{fig:vis_simulation}
        \vspace{-3pt}
\end{figure*}

\section{Implementation Details of AffordanceNet}
Beyond our reasoning-based affordance segmentation data, we also incorporate a variety of generic segmentation datasets into our training. This diverse generic set includes data for semantic segmentation (\eg, ADE20k~\cite{ade20k}, COCO-Stuff~\cite{cocostuff}, PACO~\cite{paco}), referring segmentation (\eg, RefCOCO~\cite{refcoco}), VQA (\eg, LLaVA-150k~\cite{llava}) and reasoning-based segmentation (\eg, ReasonSeg~\cite{lisa}).
The data sampling ratios are presented in Table~\ref{tab:data_ratio}.
We deploy eight NVIDIA A100 GPUs (80GB) to train our model, with a learning rate of  2e-5. 
The training loss follows~\cite{lisa}, which uses binary focal loss for map prediction and cross-entropy loss for text output.
We utilize a batch size of 40 without gradient accumulation.
\begin{table*}[h]
    \centering
    \resizebox{0.98\linewidth}{!}{
		\setlength\tabcolsep{4pt}
		\renewcommand\arraystretch{1}
    \begin{tabular}{c|cccccc}
    \toprule
        Data &  Semantic Seg & Referring Seg & Reasoning-based Seg & VQA & Affordance Seg & Reasoning-based Affordance Seg  \\
        \midrule
        Ratio &  3 & 1 & 1 & 1 & 9 & 3 \\
        \bottomrule
    \end{tabular}}
    \caption{\textbf{Data sampling ratios during training.}}
    \label{tab:data_ratio}
\end{table*}

\section{More Results on Visual Affordance}

\begin{figure*}
    \centering
    \includegraphics[width=\linewidth]{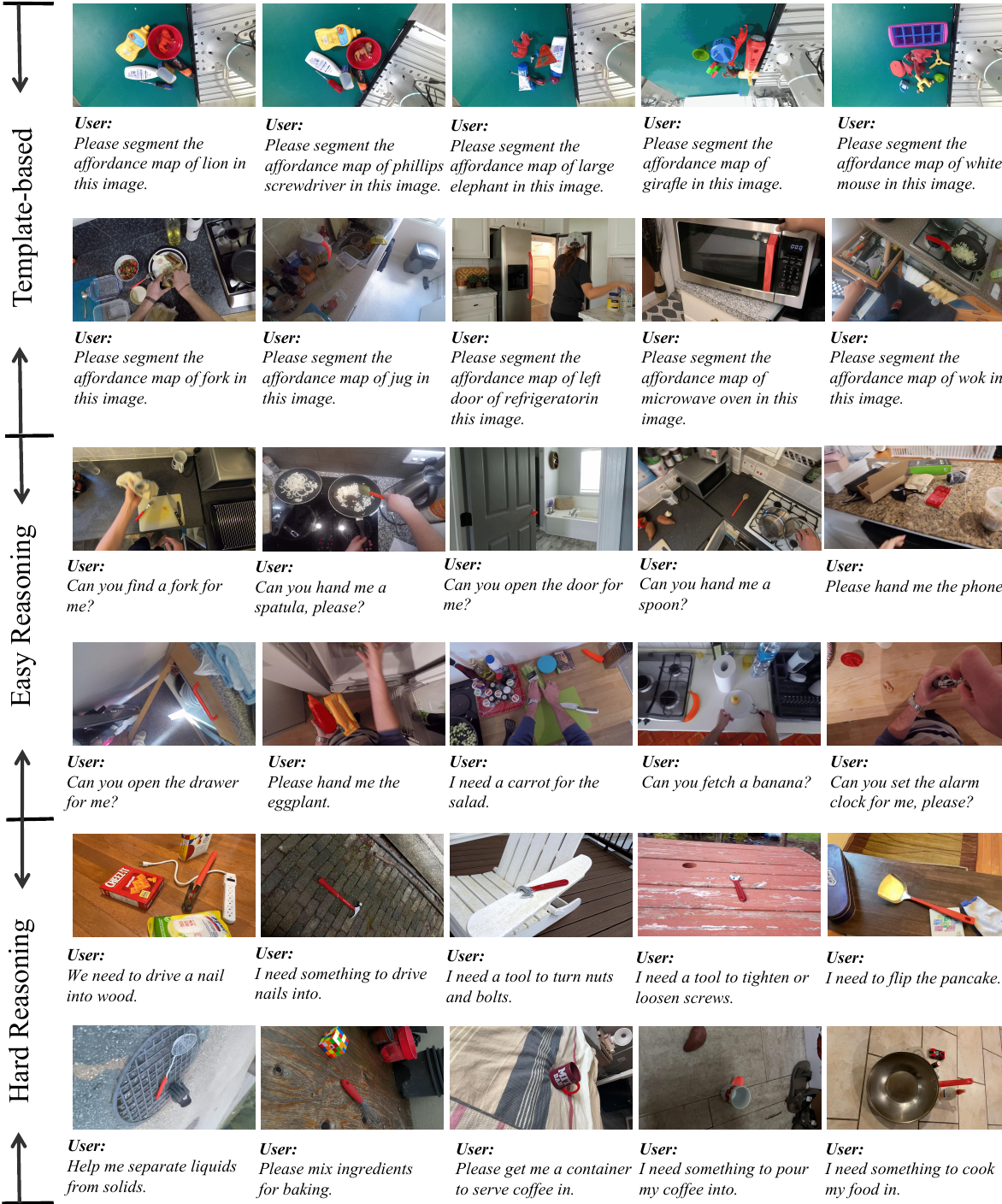}
    % \vspace{-20pt}
    \caption{\textbf{More experiment results  from our model.} We use template-based, easy reasoning-based, and hard reasoning-based instructions, respectively.}
    \label{fig:supp_results}
        \vspace{20pt}
\end{figure*}

We provide more visualization results of affordance segmentation from our AffordanceVLM model in Fig.~\ref{fig:supp_results}.
The testing images are selected from multiple validation sets, such as GraspNet \texttt{Novel}, 3DOI, and HANDAL.
We employ template-based, easy reasoning-based, and hard reasoning-based instructions for affordance map prediction, respectively.
It is obvious that our AffordanceVLM can understand these high-level human instructions, and transform them into precise affordance maps.
Meanwhile, our model can deal with various challenging situations like unseen categories or domains.
Both suggest that our model possesses robust open-world reasoning capabilities, which will significantly enhance subsequent object-grasping tasks.

\section{More Results on Real Robot}

Beyond the evaluation tasks in our main manuscript, such as grasping can, pen, screwdriver, hammer, and wok, we also evaluate the open-world generalization capabilities of our model by utilizing a broader range of instructions encompassing various unseen categories in real-robot environments, like panda, toy, circle and so on.
The real-robot experiment videos are included within the same directory.
These results demonstrate impressive open-world object perception and grasping proficiency.

\section{More Results on RLBench}

We present several visualization results from the simulation task RLBench in Figure~\ref{fig:vis_simulation}. The top video demonstrates the task ``open the top drawer," while the bottom video illustrates ``close the green jar". As shown, our model successfully completes both tasks with high accuracy.

\begin{table*}[h]
\centering
\resizebox{\linewidth}{!}{
		\setlength\tabcolsep{4pt}
		\renewcommand\arraystretch{1}
\begin{tabular}{p{0.1\textwidth}|c|p{0.8\textwidth}}
    \toprule
    \midrule
    % \multicolumn{1}{l|}{\multirow{2}*{Method}} & \multicolumn{2}{c|}{HANDAL (easy)}  & \multicolumn{2}{c|}{HANDAL (hard)} & \multicolumn{2}{c}{\cellcolor{gray!30}3DOI}\\
    % & gIoU & cIoU & gIoU & cIoU & \cellcolor{gray!30}gIoU & \cellcolor{gray!30}cIoU  \\
    Dataset & Subset & Annotation Tool and Categories \\
    \midrule
    HANDAL & - & \ding{182}: strainer, fixed joint plier, hammer, ladle, whisk, measuring cup, locking plier, power drill, adjustable wrencher, mug, ratchet, utensil, combinational wrench, pots pan, spatula, screwdriver, slip joint plier \\
    \midrule
    Open-X & RT-1 &  \ding{184}:  redbull can, rxbar blueberry, green can, apple, orange can, 7up can, sponge, pepsi can, orange, paper bowl, green rice chip bag, banana, coke can, blue chip bag, water bottle, white bowl,  rxbar chocolate, 7up can, brown chip bag, blue plastic bottle, green jalapeno chip bag, blue water bottle, \\
    & &\ding{186}: right fridge door,  bottom drawer,  left fridge door, middle drawer, top drawer  \\
    % \hline
    \cmidrule(lr){2-3} % 覆盖第2列到第3列
    & Bridge & \ding{184}: apple, apple slice, avocado, ball, banana, banana plush, baster, beet, beetroot,  bell pepper, berry, blackberry, board, book, bot, bottle, bowel, bowl, bread, bread roll, broccoli, bunny, butter, cake, cake slice, can, cap, capsicum, carrot, cereal, cheese, cheese slice, cheese wedge, cherry, cake, cake slice, can, cap, capsicum, carrot, cauliflower, cereal, cheese slice, cherry, chicken drumstick, chicken leg, chicken piece, chili pepper, chocolate, croissant, cucumber, detergent, dishcloth, doll, dough, drumstick, egg, eggplant, eggroll, garlic, half bun, hot dog, hotdog, lime, lobster tail, mango, meat, mouse, plastic fish, plush animal, sausage, soap, stuffed animal, stuffed dog, stuffed mushroom, stuffed cheetah, stuffed duck, stuffed pig, strawberry, sushi, tube, turkey leg, yam\\
    && \ding{185}: knife\\
    && \ding{186}: brush, cutter, drawer of box, fork, gripper, hairbrush, ice cream scoop, kettle, laddle, microwave, mug, oven, pot, pan, saucepan, scissors, scrub brush, scrubber, spatula, spork, teapot, teal brush, wok\\
    \midrule
    EgoObjects &-&\ding{183}: alarm clock, balloon, blanket, book, bottle, bowl, box, computer mouse, doll, envelope, eraser, flowerpot, flying disc, football, game controller/pad, glasses, glove, goggles, lipstick, necklace, paper, paper towel, pen, pencil, pencil case, perfume, phone charger, picture frame, pillow, plate, post-it, poster, pottery, remote control, ring, shirt, shorts, skateboard, soap, sock, stapler, sun hat, sunglasses, tablet computer, teddy bear, tennis ball, toothpaste, towel, umbrella, vase, wallet, watch \\
    && \ding{185}: spoon, mug, screwdriver, knife, wrench \\
    && \ding{186}: microwave oven, washing machine, wok, oven, drawer, teapot, toothbrush, wardrobe, door, jug, refrigerator, tap, tennis racket, spatula, fork, frying pan, scissors, hammer\\
    \midrule
    GraspNet & - & \ding{182}: dish, cracker box, pear, camel, peach, tape, banana, head shoulders care, black mouse, tomato soup can, darlie toothpaste, rhinocero, baoke marker, hosjam, pantene, racquetball, cups, sum37 secret repair, gorilla, kispa cleanser, hippo, toy airplane, dabao wash soup, weiquan, strawberry, dabao facewash, head shoulders supreme, dabao sod, large elephant, darlie box, nzskincare mouth rinse, plum \\
   && \ding{186}:   ﬂat screwdriver, power drill, scissors, mug\\
       \midrule
       \midrule
    HANDAL \textit{Reasoning} & - & \textbf{Hard Instructions}: strainer, fixed joint plier, hammer, ladle, whisk, measuring cup, locking plier, power drill, adjustable wrencher, mug, ratchet, utensil, combinational wrench, pots pan, spatula, screwdriver, slip joint plier \\
       \midrule
    EgoObjects \textit{Reasoning}  &-& \textbf{Easy Instructions}:  alarm clock, balloon, blanket, book, bottle, bowl, box, computer mouse, doll, envelope, eraser, flowerpot, flying disc, football, game controller/pad, glasses, glove, goggles, lipstick, necklace, paper, paper towel, pen, pencil, pencil case, perfume, phone charger, picture frame, pillow, plate, post-it, poster, pottery, remote control, ring, shirt, shorts, skateboard, soap, sock, stapler, sun hat, sunglasses, tablet computer, teddy bear, tennis ball, toothpaste, towel, umbrella, vase, wallet, watch, spoon, mug, screwdriver, knife, wrench, microwave oven, washing machine, wok, oven, drawer, teapot, toothbrush, wardrobe, door, jug, refrigerator, tap, tennis racket, spatula, fork, frying pan, scissors, hammer\\
    && \textbf{Hard Instructions}: spoon, mug, screwdriver, knife, wrench, microwave oven, washing machine, wok, oven, drawer, teapot, toothbrush, wardrobe, jug, refrigerator, tap, tennis racket, spatula, fork, frying pan, scissors, hammer \\
    \midrule
    \bottomrule
\end{tabular}
}
\caption{\textbf{Annotation details of RAGNet on tool composition} for different subsets and categories.}
\vspace{60pt}
\label{tbl:supp_train_data}
\end{table*}

\begin{figure*}
    \centering
    \includegraphics[width=\linewidth]{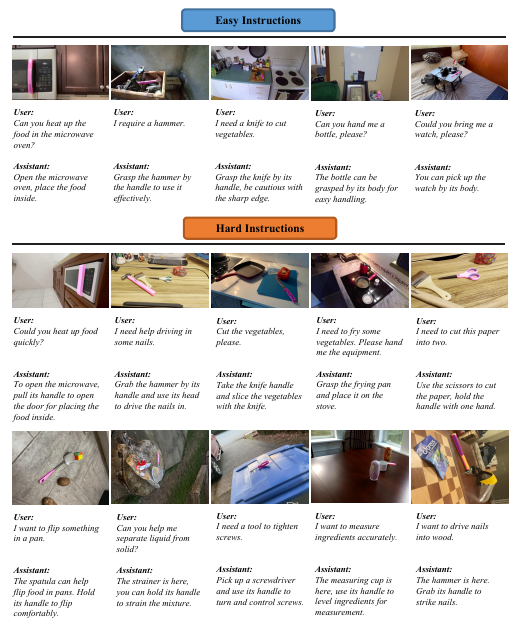}
    \vspace{-20pt}
    \caption{\textbf{More reasoning-based affordance segmentation examples from our RAGNet.} It includes two types of reasoning instructions: easy instructions and hard instructions.}
    \label{fig:supp_reasoning}
\end{figure*}

\begin{table*}[h]\centering
\begin{minipage}{0.9\linewidth}\vspace{0mm}    \centering
\begin{tcolorbox} 
    \small
     \hspace{-6mm}
\textcolor{blue}{messages} = [`role': `system', `content': `You are a helpful assistant. Based on several words where the first is category name, please design an instruction $<1>$ and instruction $<2>$ in embodied scenes. The instruction $<1>$ must include object category name itself. The instruction $<2>$ must include object category name itself. The instruction $<2>$ must belongs to embodied manipulation and give action if instruction  $<1>$  provides. The instruction $<2>$does not exceed 50 words.', \\
`role': `user', `content': `mug',\\
`role': `assistant', `content': `$<1>$ I need a drink. Please find a mug to fill water. $<2>$ The mug has a handle as affordance map. So the robot can hold its handle.' \\
`role': `user', `content': `knife' \\
`role': `assistant',
`content': `$<1>$ Please give me a knife to cut apple. $<2>$ The knife has a handle, and you can use its handle to cut apple.', \\
`role': `user', `content': `hammer',\\
`role': `assistant', `content': `$<1>$  What is the proper way to hold the hammer? 
 $<2>$ The correct method is to hold the hammer by its handle.', \\
`role': `user', `content': `fork',\\
`role': `assistant', `content': $<1>$  Kindly pick up the fork. $<2>$ You will be holding the fork handle.', \\
`role': `user', `content': `screwdriver',\\
`role': `assistant', `content': `$<1>$  I need a tool to tighten or loosen screws. $<2>$ The screwdriver is here, hold its handle to turn and control screws.', \\
`role': `user', `content': `\textcolor{red}{\texttt{words}}'
]
\end{tcolorbox}
% \vspace{-2mm}
\caption{Language prompt when \textbf{generating easy reasoning-based instructions on HANDAL via GPT-4.}}
    \label{table:supp_easy_reasoning}
\end{minipage}
\vspace{2mm}
\end{table*}

\begin{table*}[h]\centering
\begin{minipage}{0.9\linewidth}\vspace{0mm}    \centering
\begin{tcolorbox} 
    \small
     \hspace{-6mm}
\textcolor{blue}{messages} = [`role': `system', `content': `You are a helpful assistant. Based on several words where the first is category name, please design an instruction $<1>$ and instruction $<2>$ in embodied scenes. The instruction $<1>$ must not include object category name itself. The instruction $<2>$ must include object category name itself. The instruction $<2>$ must belongs to embodied manipulation and give action if instruction  $<1>$  provides. The instruction $<2>$does not exceed 50 words.', \\
`role': `user', `content': `microwave, open',\\
`role': `assistant', `content': `$<1>$ Heat up food quickly . $<2>$ The microwave is closed, so it can be open to access the food inside.' \\
`role': `user', `content': `knife' \\
`role': `assistant',
`content': `$<1>$ I want to cut a bread. $<2>$ The knife has a handle, you can use its handle to cut bread.', \\
`role': `user', `content': `computer mouse',\\
`role': `assistant', `content': `$<1>$  Give me a tool to control the cursor on the screen. $<2>$ The computer mouse is here. It has not handle, so you can grasp its whole body.', \\
`role': `user', `content': `fork',\\
`role': `assistant', `content': $<1>$  Use to pierce and lift food. $<2>$ The fork is here, and its handle can be grasped.', \\
`role': `user', `content': `screwdriver',\\
`role': `assistant', `content': `$<1>$ I need a tool to tighten or loosen screws. $<2>$ The screwdriver is here, hold its handle to turn and control screws.', \\
`role': `user', `content': `\textcolor{red}{\texttt{words}}'
]
\end{tcolorbox}
\vspace{-2mm}
\caption{Language prompt when \textbf{generating hard reasoning-based instructions on HANDAL via GPT-4.}}
    \label{table:supp_hard_reasoning}
\end{minipage}
\vspace{30pt}
\end{table*}

\end{document}